\documentclass[letterpaper, 10 pt, journal, twoside]{IEEEtran}

\usepackage{amsthm}

\usepackage{amsmath}  
\usepackage{amssymb}  
\usepackage{amsfonts}
\usepackage{mathtools}
\usepackage{bm}
\DeclareMathAlphabet{\pazocal}{OMS}{zplm}{m}{n}

\usepackage{csquotes}
\usepackage{paralist}

\usepackage{verbatim}

\usepackage{color}
\usepackage{graphicx}
\usepackage{adjustbox}

\usepackage{mathptmx}
\usepackage{times}
\usepackage[hang,flushmargin]{footmisc}
\usepackage{multirow}

\usepackage[ruled,vlined,linesnumbered]{algorithm2e}
\usepackage[noend]{algorithmic}
\SetKwInput{KwInput}{Input}                
\SetKwInput{KwOutput}{Output}              

\usepackage[us]{datetime}
\usepackage{caption}
\usepackage[normalem]{ulem}

\usepackage[usenames,dvipsnames,table]{xcolor}

\usepackage{hyperref} 
\hypersetup{
  colorlinks,
  citecolor=black,
  filecolor=black,
  linkcolor=black,
  urlcolor=black,
  pdfauthor={},
  pdfsubject={},
  pdftitle={}
}

\newcommand{\Hl}[2][\empty]{%
\ifx#1\empty
\else
\sethlcolor{#1}%
\fi
\hl{#2}}
\usepackage{soul}
\soulregister\Hl{7}
\soulregister\ref7
\soulregister\autoref7
\soulregister\eqref7
\soulregister\cite7
\soulregister\pageref7
\soulregister\ac7
\soulregister\acp7
\soulregister\acs7
\soulregister\acl7
\soulregister\subsection7
\soulregister\SI7
\soulregister\url7

\newcommand{\rev}[1]{#1} 
\usepackage{mdframed}

\usepackage{todonotes}

\usepackage{subcaption}

\usepackage{latexsym}
\usepackage{color}

\usepackage{cite}

\usepackage{siunitx}
\usepackage{ifthen}

\usepackage{booktabs}
\usepackage{tablefootnote}

\usepackage{subfiles}
\usepackage{bm}
\usepackage{titlesec}

\titlespacing*{\section}{0pt}{6pt}{1pt}
\titlespacing*{\subsection}{0pt}{4pt}{1pt}

\usepackage{pifont}
\definecolor{color_green}{rgb}{0, .722, .243}
\definecolor{color_red}{rgb}{0.737,0.165,0}

\usepackage{stackengine} 

\usepackage{tikz}
\usetikzlibrary{calc}
\usepackage{tikz-dimline}
\usepackage{physics}
\usepackage{tikz-3dplot}
\usepackage[outline]{contour} 
\usetikzlibrary{fadings}
\usetikzlibrary{decorations.markings}

\tikzfading[name=fade up,top color=red!100, bottom color=white!0]

\tikzset{testfade/.style n args={3}{
    postaction={
    decorate,
    decoration={
    markings,
    mark=between positions 0 and \pgfdecoratedpathlength step 0.5pt with {
    \pgfmathsetmacro\myval{multiply(
        divide(
        \pgfkeysvalueof{/pgf/decoration/mark info/distance from start}, \pgfdecoratedpathlength
        ),
        100
    )};
    \pgfsetfillcolor{#3!\myval!#2};
    \pgfpathcircle{\pgfpointorigin}{#1};
    \pgfusepath{fill};}
}}}}

\colorlet{veccol}{green!45!black}
\colorlet{myred}{red!80!black}
\colorlet{myblue}{blue!80!black}
\colorlet{mypurple}{blue!50!red!100!}
\colorlet{amber}{red!90!yellow!100!blue!50}
\colorlet{mygreen}{green!40!black}
\colorlet{projcol}{blue!70!black}
\colorlet{mydarkblue}{blue!50!black}
\colorlet{veccol}{green!50!black}
\usetikzlibrary{shapes.geometric,arrows.meta,angles,quotes}
\tikzstyle{thin arrow}=[dashed,thin,-{Latex[length=4,width=3]}]
\tikzstyle{arrow}=[{-Latex}]

\tikzset{>=latex} 
\tikzstyle{proj}=[projcol!80,line width=0.08] 
\tikzstyle{area}=[draw=veccol,fill=veccol!80,fill opacity=0.6]
\tikzstyle{vector}=[-stealth,myblue,thick,line cap=round]
\tikzstyle{unit vector}=[->,veccol,thick,line cap=round]
\tikzstyle{dark unit vector}=[unit vector,veccol!70!black]
\usetikzlibrary{angles,quotes} 
\contourlength{1.3pt}

\definecolor{color_blue}{rgb}{0.22, 0.2, 0.502}
\definecolor{color_red}{rgb}{0.737,0.165,0}
\definecolor{color_green}{rgb}{0, .522, .243}

\usepgfplotslibrary{fillbetween}
\usepgfplotslibrary{groupplots}
\usepgfplotslibrary{colormaps} 
\def\centerarc[#1](#2)(#3:#4:#5)
    { \draw[#1] ($(#2)+({#5*cos(#3)},{#5*sin(#3)})$) arc (#3:#4:#5); }

\DeclareMathOperator*{\minimize}{minimize}

\usepackage{calrsfs}
\DeclareMathAlphabet{\pazocal}{OMS}{zplm}{m}{n}

\newcommand{\position}{p}
\newcommand{\positionvector}{\bm{\position}}
\newcommand{\quaternion}{q}
\newcommand{\orientation}{\bm{\quaternion}}
\newcommand{\linearvel}{v}
\newcommand{\linearvelvector}{\bm{\linearvel}}
\newcommand{\angularvel}{\omega}
\newcommand{\angularvelvector}{\bm{\angularvel}}

\newcommand{\statestd}{\bm{x}}

\newcommand{\motorforce}{f}

\newcommand{\bodythrustvector}{\bm{f}_{T}}

\newtheoremstyle{remark}%
  {3pt}
  {3pt}
  {\itshape}
  {}
  {\bfseries}
  {}
  {.2em}
  {\thmname{#1}: \thmnote{\normalfont#3}}

\theoremstyle{remark}
\newtheorem*{remark}{Remark}

\usepackage[printonlyused]{acronym}
    \acrodef{uav}[UAV]{Uncrewed Aerial Vehicle}
    \acrodef{ros}[ROS]{Robot Operating System}
    \acrodef{nmpc}[NMPC]{Nonlinear Model Predictive Control}
    \acrodef{mpc}[MPC]{Model Predictive Control}
    \acrodef{imu}[IMU]{Inertial Measurement Unit}
    \acrodef{rvo}[RVO]{Reciprocal Velocity Obstacle}
    \acrodef{rvc}[RVC]{Reciprocal Velocity Constraint}
    \acrodef{vo}[VO]{Velocity Obstacle}
    \acrodef{orca}[ORCA]{Optimal Reciprocal Collision Avoidance}
    
\usepackage[placement=top,vshift=1,firstpage=true]{background}\SetBgScale{1.0}
\SetBgContents{\parbox{0.95\textwidth}{\small \begin{center} \copyright{} 2026 The Authors. This work is licensed under a Creative Commons Attribution 4.0 License. For more information, see https://creativecommons.org/licenses/by/4.0/. This work has been accepted for publication in IEEE Robotics and Automation Letters. \end{center}}}
\SetBgColor{black}
\SetBgAngle{0}
\SetBgOpacity{1.0}

\title{%
\Title
}

\title{\LARGE \bf RVC-NMPC: Nonlinear Model Predictive Control with Reciprocal Velocity Constraints for Mutual Collision Avoidance\\ in Agile UAV Flight}

\author{V\'{i}t Kr\'{a}tk\'{y}$^{1}$, Robert P\v{e}ni\v{c}ka$^1$, Parakh M. Gupta$^1$, Ond\v{r}ej Proch\'{a}zka$^1$, and Martin Saska$^1$
    \thanks{Manuscript received: May, 31, 2026; Revised July, 29, 2026; Accepted August, 23, 2026.}
    \thanks{This paper was recommended for publication by Editor Aniket Bera upon evaluation of the Associate Editor and Reviewers' comments. This work was supported by the Czech Science Foundation (GA\v{C}R) under research project no. 23-06162M, by the European Union under the project Robotics and advanced industrial production (reg.no. CZ.02.01.01/00/22 008/0004590), and by the CTU grant no. SGS23/177/OHK3/3T/13.}
    \thanks{$^1$Authors are with the Department of Cybernetics, Faculty of Electrical Engineering, Czech Technical University in Prague, Czech Republic, {\tt \footnotesize \href{mailto:vit.kratky@fel.cvut.cz}{vit.kratky}|\href{mailto:robert.penicka@fel.cvut.cz}{robert.penicka}|\href{mailto:guptapar@fel.cvut.cz}{guptapar}|\href{mailto:prochon4@fel.cvut.cz}{prochon4}|
    \href{mailto:martin.saska@fel.cvut.cz}{martin.saska}\}@fel.cvut.cz}}
\thanks{Digital Object Identifier (DOI): 10.1109/LRA.2026.3737070.}
}

\begin{document}
\maketitle
\setcounter{footnote}{0} 

\begin{abstract}
  This paper presents an approach to mutual collision avoidance based on \ac{nmpc} with time-dependent \acp{rvc}.
  Unlike most existing methods, the proposed approach relies solely on observable information about other robots, eliminating the need for excessive communication.   
  The computationally efficient algorithm for computing \acp{rvc}, together with the direct integration of these constraints into the \ac{nmpc} problem formulation at the controller level, allows the whole pipeline to run at \SI{100}{\hertz}. 
  This high processing rate, combined with modeled nonlinear dynamics of the controlled \acp{uav}, is a key feature that facilitates the use of the proposed approach for agile \ac{uav} flight.      
  The proposed approach was evaluated through extensive simulations emulating real-world conditions in scenarios involving up to 10 \acp{uav} and velocities of up to \SI{25}{\meter\per\second}, and in real-world experiments with accelerations up to \SI{30}{\meter\per\second\squared}.
  Comparison with the state of the art shows a \SI{31}{\percent} improvement in terms of flight time reduction in challenging scenarios, while maintaining collision-free navigation in all trials.
  
\end{abstract}



\IEEEpeerreviewmaketitle

\section*{Supplementary material}

\textbf{Video:} {\small\url{https://youtu.be/LYnn-eDvkec}} 


\section{Introduction}
\label{sec:introduction}

\IEEEPARstart{T}{he} deployment of \aclp{uav} (\acsp{uav}) over the past decade has been mostly limited to single-robot applications in isolated operational spaces. However, in recent years, development has targeted applications with numerous \acp{uav} operating in an open-air space shared with other air-traffic participants (e.g., package delivery and area monitoring).   
This brings to the forefront the problem of mutual collision avoidance, a key aspect of the safe deployment of robotic systems in real-world applications where robots share the operational space.

Once \acp{uav} are deployed on an everyday basis for a great variety of tasks, they are expected to operate in much denser environments than, e.g., airplanes due to their limited flight altitudes and higher density of starting and delivery locations. 
Under such conditions, centralized planning and scheduling become impractical. 
Consequently, decentralized methods that enable reliable collision avoidance during high-speed, agile flight are of particular importance, as they allow \acp{uav} to fully exploit their efficiency and maneuverability.

Approaches addressing mutual collision avoidance in multi-robot scenarios mostly focus on providing theoretical guarantees but neglect real-world aspects of the problem~\cite{cheng2017DecentralizedNavigationMultiple, vandenberg2010orca, levy2015ExtendedVelocityObstacle, guo2021VRORCAVariableResponsibility, toumieh2024motionPlanning, tordesillas2022MADERTrajectoryPlannera, DLSCDistributedMultiAgent}.  
Most of these works assume unrealistic perfect control (reference tracking)~\cite{cheng2017DecentralizedNavigationMultiple}, often neglect the kinematic and dynamic constraints of the \acp{uav}~\cite{vandenberg2010orca, levy2015ExtendedVelocityObstacle, guo2021VRORCAVariableResponsibility} or require knowledge of the future trajectories of all other \acp{uav} which puts high requirements on the communication network bandwidth~\cite{toumieh2024motionPlanning, tordesillas2022MADERTrajectoryPlannera, DLSCDistributedMultiAgent}.  
Even under these unrealistic or highly restrictive assumptions, most existing works are unable to handle scenarios involving velocities exceeding \SI{10}{\meter\per\second}, which is well below the speeds achievable by commercially available drones\footnote{https://enterprise.dji.com/matrice-30/specs, https://www.skydio.com/x10}.  
\begin{figure}[tb]
\centering
  \includegraphics[width=\columnwidth]{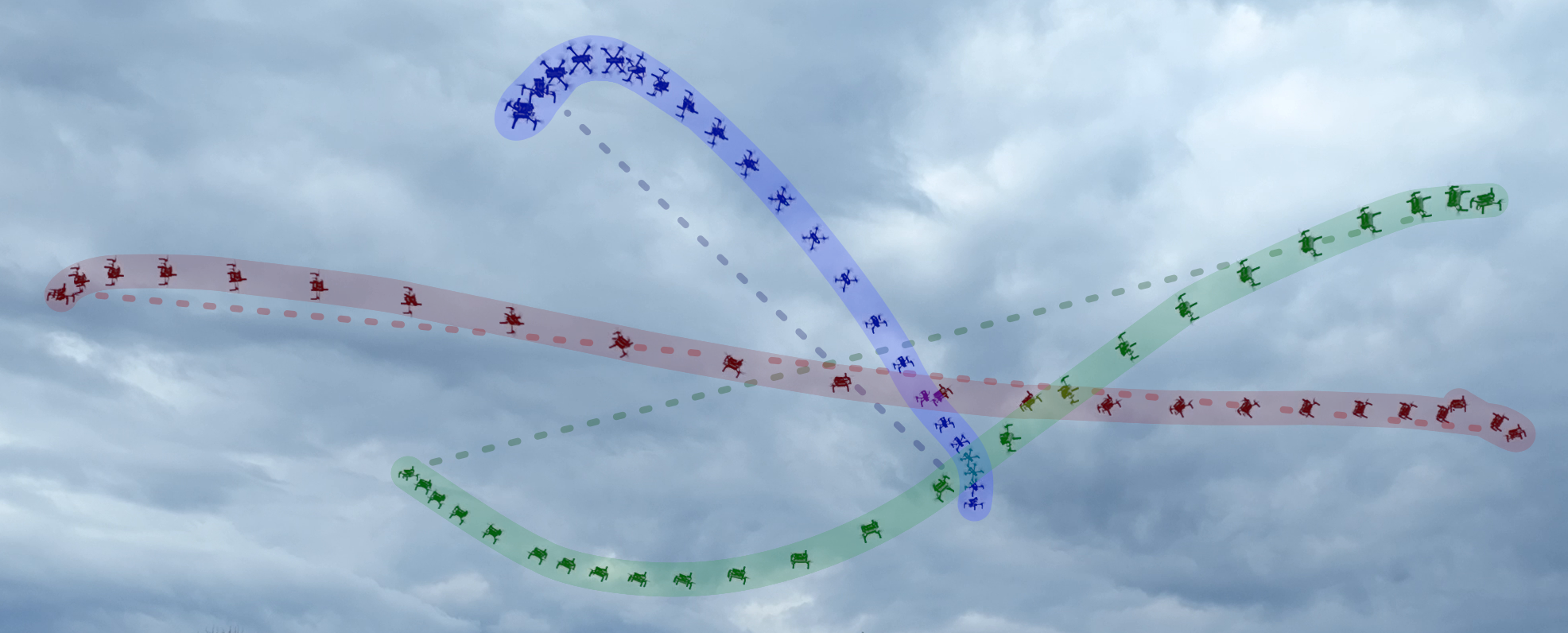}
  \caption{Deployment of the introduced RVC-NMPC approach in a real-world scenario with 3 UAVs navigating to antipodal positions on a circle with radius \SI{10}{\meter}, and acceleration limit \SI{30}{\meter\per\second\squared}.}
  \vspace*{-0.3cm}
\label{fig:intro}
\end{figure}

To this end, we address the problem of mutual collision avoidance by proposing a novel \ac{nmpc}-based approach with time-dependent \aclp{rvc} (\acsp{rvc}) that are computed only based on the current position and velocity of the robots. 
In contrast to future trajectories required to be communicated between robots by state-of-the-art methods, the position and velocity can be obtained by other \acp{uav} through onboard sensing~\cite{xu2023onboardDetection, vrba2024lidarDetection} or through a low-bandwidth communication network, e.g., as part of the Remote Drone ID\footnote{https://drone-remote-id.com/}.     
Integrating \acp{rvc} directly into the \acp{uav} control pipeline ensures proper and fast reaction to external disturbances, increases the method's reliability, and allows seamless integration of dynamic constraints. 
Low computational demands enable all pipeline modules to run at \SI{100}{\hertz} on a \SI{2}{\giga\hertz} ARM processor, further facilitating fast reaction to changes in the behavior of other \acp{uav}, thus enabling the efficient use of the method in high-speed scenarios. 

Despite the absence of theoretical guarantees, the proposed RVC-NMPC approach demonstrated its practicality through collision-free navigation in a 3-hour-long test in a simulation with 10 robots following trajectories with velocities and accelerations up to \SI{25}{\meter\per\second}, and \SI{40}{\meter\per\second\squared}, respectively, and also through real-world experiments with three \acp{uav} navigating with velocities up to \SI{18}{\meter\per\second} and accelerations up to \SI{30}{\meter\per\second\squared}. 
The approach further shows superior performance in scenarios maximizing the number of potential collisions in obstacle-free environments, where it reduces the average time needed for all robots to reach their goals by \SI{31}{\percent} compared to state of the art while not experiencing any collision.
Additional analyses are provided to demonstrate the method's robustness with respect to communication delays and noise in the estimation of states of other \acp{uav}.

\section{Related Work}

\label{sec:relatedWork}

The problem of finding collision-free trajectories in multi-robot scenarios can be solved both in a centralized and decentralized manner.
In this review of related work, we omit centralized approaches~\cite{augugliaro2012globalplanning, honig2018planning, solovey2016planning} in favor of decentralized solutions, as the centralized approaches are impractical for high-speed agile flight due to additional communication delays and poor scalability.  

In recent years, the major focus has been on the development of decentralized methods that require the robots to share not only the current state of \acp{uav}, but also their planned trajectories.
The majority of these methods rely on optimization techniques and vary in trajectory parametrization, and methods for planning, free space decomposition, and obstacle representations~\cite{zhou2021egoswarm, swarmingInTheWild, hou2022egoEnhanced, tordesillas2022MADERTrajectoryPlannera, kondo2024RobustMADERDecentralized, kondo2024PUMAFullyDecentralized, DLSCDistributedMultiAgent,toumieh2024motionPlanning}. 
Some of these works address individual aspects of the cooperative navigation problem, such as efficient collision resolution in dense environments~\cite{hou2022egoEnhanced}, robustness to communication delays~\cite{kondo2024RobustMADERDecentralized}, perception- and uncertainty-awareness~\cite{kondo2024PUMAFullyDecentralized}, or deadlock prevention~\cite{DLSCDistributedMultiAgent}.
However, while these works often focus on fast navigation and measure their performance as time required to task completion, the maximum achieved and allowed velocities are mostly limited to a few meters per second. 
A superior approach considering high-speed navigation in multi-agent scenarios was introduced in~\cite{toumieh2024motionPlanning}, where the HDSM algorithm is demonstrated to navigate complex scenarios with average speeds of up to \SI{3.61}{\meter\per\second}, \SI{100}{\percent} success rate, and theoretical guarantees.
While these methods show impressive results in cluttered environments, they require complete knowledge of other \acp{uav}' states and their planned trajectories. 
This limits their use to scenarios with cooperating robots sharing the required data and puts high demands on communication network bandwidth.

Reactive approaches for mutual collision avoidance often rely on representing other robots as obstacles with simplified dynamics (e.g., \ac{vo}~\cite{fiorini1998velocityobstacles}).
The most direct extensions of \ac{vo} concept are \acp{rvo}~\cite{vandenberg2008rvo}, \ac{orca}~\cite{vandenberg2010orca}, and V-RVO~\cite{arul2021vrvo} which improves the efficiency of collision avoidance by letting each agent take half of the responsibility to avoid collision between cooperating robots. 
While these approaches implement collision avoidance based on position and velocity observations only, they neglect physical constraints of individual platforms and assume an immediate change of their velocity.

The lack of consideration of dynamic models is overcome in several adaptations of velocity obstacles, e.g., by using second-order dynamics~\cite{vandenberg2011avo}, nonholonomic models~\cite{javier2012nonholonomicrvo}, and general linear systems~\cite{bareiss2013qrobstacles, vandenberg2012lqg}.
Some approaches overcome the simplicity of these concepts by integrating \acp{vo} or their adaptations with other frameworks such as reinforcement learning~\cite{qin2024SRLORCASociallyAware, liu2023MAPPOBasedOptimalReciprocal} or \ac{mpc}~\cite{cheng2017DecentralizedNavigationMultiple, arul2020DCADDecentralizedCollision}. 
In~\cite{cheng2017DecentralizedNavigationMultiple}, the authors use \ac{orca} constraints directly in a \ac{mpc} formulation of the trajectory generation problem to account for physical constraints of the robots. 
The DCAD approach introduced in~\cite{arul2020DCADDecentralizedCollision} further develops this idea by integrating the downwash in the \ac{orca} algorithm and addressing nonlinearities through flatness-based feedforward linearization.  
However, these methods formulate collision-avoidance constraints for every transition point along the prediction horizon, which substantially increases computational complexity. Consequently, they rely on simplified models, shorter prediction horizons, or lower update rates to maintain real-time feasibility, limiting their applicability in high-speed agile flight.

Beyond MPC-based approaches, the use of \ac{nmpc} for mutual collision avoidance has also been explored in the literature~\cite{zhu2019ChanceConstrainedCollisionAvoidance, goarin2024DecentralizedNonlinearModel}.
While \ac{nmpc} addresses system nonlinearity and offers greater flexibility for integrating collision avoidance mechanisms directly into the control problem formulation, it is prone to a significant increase in computational demands due to extensive use of nonlinear constraints and overly complex problem formulation, which limits the prediction horizon length and update rate, thereby restricting the applicability of these methods in high-speed scenarios.

While the majority of related works either require knowledge of planned trajectories of other \acp{uav} or neglect the physical constraints of the robots, none of these works has demonstrated reliable navigation in high-speed scenarios exceeding \SI{10}{\meter\per\second}.
In contrast to existing ORCA-MPC approaches, the proposed RVC-NMPC method computes a single set of reciprocal velocity constraints from the currently observed states of neighboring robots and applies them over the entire prediction horizon. This substantially reduces the number of constraints that must be generated and applied within the NMPC problem, enabling operation at 100 Hz while preserving nonlinear dynamic modeling and requiring knowledge of the current positions and velocities of other robots only.
These properties distinguish the proposed RVC-NMPC approach from the state of the art and, together with the consideration of kinematic and dynamic constraints and the emphasis on maintaining feasibility of the defined NMPC problem, make the method suitable for high-speed agile flight in real-world environments without imposing any unrealistic assumptions.



\vspace{-0.2cm}
\section{Preliminaries}
\label{sec:preliminaries}

The proposed approach is not limited to a particular type of aerial vehicle or dynamic model. However, in the remainder of this manuscript, we use standard quadrotor dynamic model equivalent to that specified in~\cite{gupta2025lolnmpclowleveldynamicsintegration}. 
\rev{The quadrotor's state is represented by $\statestd=\begin{bmatrix} \positionvector,\orientation,\linearvelvector,\angularvelvector\end{bmatrix}^{T}$ which comprises of position $\positionvector \in \mathbb{R}^3$, velocity $\linearvelvector \in \mathbb{R}^3$, unit quaternion rotation $\orientation \in SO(3)$, and body rates or angular velocity of the aircraft in the body-frame $\angularvelvector \in \mathbb{R}^3$.
  The input of the model is given as a vector of single-rotor thrusts $\bm{f} = \begin{bmatrix} f_1,f_2,f_3,f_4 \end{bmatrix}$.}

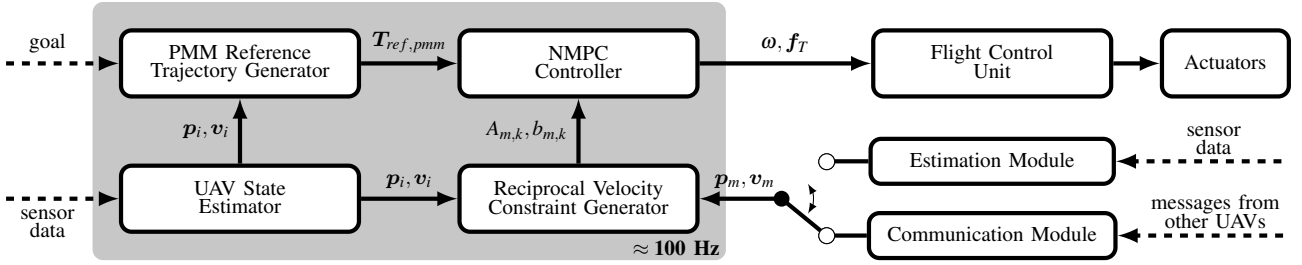
\begin{figure*}[htb]
  \centering
\pgfdeclarelayer{backbackground}
\pgfdeclarelayer{background}
\pgfdeclarelayer{foreground}
\pgfsetlayers{backbackground,background,main,foreground}

\definecolor{arrow_red}{HTML}{A30D00}
\definecolor{arrow_green}{rgb}{0, .522, .243}

\definecolor{arrow_blue}{HTML}{22559C}

\definecolor{color_grey}{HTML}{606060}
\definecolor{color_blue}{HTML}{22559C}
\definecolor{color_red}{HTML}{FF3333}
\definecolor{color_green}{HTML}{D9F8C4}

\pgfmathsetmacro{\vshift}{0.08em} 
\pgfmathsetmacro{\hshift}{0.35em} 
\pgfmathsetmacro{\hshifta}{0.33em} 
\pgfmathsetmacro{\hshiftb}{0.43em} 
\pgfmathsetmacro{\voffset}{0.09em} 
\pgfmathsetmacro{\hoffset}{0.00em} 
\pgfmathsetmacro{\folloffset}{0.325em} 
\pgfmathsetmacro{\folloffsetn}{0.12em} 
\pgfmathsetmacro{\operatoroffset}{0.055em} 
\pgfmathsetmacro{\postdeployoffset}{0.124em} 
\pgfmathsetmacro{\postdeployvoffset}{0.118em} 
\pgfmathsetmacro{\nthuavvoffset}{-0.1em} 
\pgfmathsetmacro{\predeployoffset}{-0.27em} 
\pgfmathsetmacro{\predeployvoffset}{-0.066em} 
\pgfmathsetmacro{\safetyoperatoroffset}{-0.258em} 
\pgfmathsetmacro{\legendoffset}{-0.08em} 
\pgfmathsetmacro{\sensorsshift}{0.43} 
\pgfmathsetmacro{\trackingshift}{-0.06} 
\pgfmathsetmacro{\harrowshift}{0.1} 
\pgfmathsetmacro{\harrowoffset}{0.12} 
\pgfmathsetmacro{\varrowoffset}{0.18} 
\pgfmathsetmacro{\varrowshift}{0.18} 
\pgfmathsetmacro{\harrowshiftpredeploy}{1.69} 
\pgfmathsetmacro{\harrowshiftpostdeploy}{1.4} 
\pgfmathsetmacro{\harrowshiftoperator}{0.4} 
\pgfmathsetmacro{\operatorswitchshift}{0.5} 
\pgfmathsetmacro{\harrowbend}{0.1em} 
\pgfmathsetmacro{\safetyswitchshift}{0.092em} 
\pgfmathsetmacro{\harrowshiftscan}{0.14em} 
\pgfmathsetmacro{\predeployintershift}{0.52} 
\pgfmathsetmacro{\operatorintershift}{0.25} 
\pgfmathsetmacro{\postdeployintershift}{0.165} 
\pgfmathsetmacro{\representationshift}{0.04} 

\tikzstyle{block}=[draw, rounded corners, text centered, ultra thick, minimum height=2.5em, minimum width=9.0em, inner sep=1pt, fill=white, fill opacity=1.0, text opacity=1.0]
\tikzstyle{block_act}=[draw, rounded corners, text centered, ultra thick, minimum height=2.5em, minimum width=4.8em, inner sep=1pt, fill=white, fill opacity=1.0, text opacity=1.0]
\tikzstyle{block_single}=[draw, rounded corners, text centered, ultra thick, minimum height=1.7em, minimum width=9.2em, inner sep=1pt, fill=white, fill opacity=1.0, text opacity=1.0]
\tikzstyle{block_filter}=[draw, rounded corners, text centered, minimum height=2.0em,  minimum width=5.6em, fill=white, fill opacity=1.0, text opacity=1.0]
\tikzstyle{block_perf}=[draw, rounded corners, text centered, minimum height=2.0em, minimum width=5.6em, fill=white, fill opacity=1.0, text opacity=1.0]
\def\nodedst{2.0cm}
\tikzstyle{arrow}=[draw, ->, thick]
\def\nodedst{2.0cm}

\begin{tikzpicture}[auto, node distance=1.0cm, >=latex, font=\scriptsize]

  \renewcommand{\arraystretch}{0.6}


  \begin{pgfonlayer}{foreground}


    \node [block, shift = {(0.0cm, 0.0cm)}] (pmm) {
        \begin{tabular}{c}
          \footnotesize PMM Reference \\
          \footnotesize Trajectory Generator\\
    \end{tabular}};

    \node [block, right of=pmm, shift = {(\hshift, 0.0cm)}] (nmpc) {
        \begin{tabular}{c}
          \footnotesize NMPC \\
          \footnotesize Controller \\
    \end{tabular}};

    \node [block, right of=nmpc, shift = {(\hshift+1.0, 0.0cm)}] (fcu) {
        \begin{tabular}{c}
          \footnotesize Flight Control \\
          \footnotesize Unit \\
    \end{tabular}};

    \node [block_act, right of=fcu, shift = {(0.6*\hshift, 0.0cm)}] (actuators) {
        \begin{tabular}{c}
          \footnotesize Actuators \\
    \end{tabular}};

    \node [block, below of=pmm, shift = {(-0.0cm, -\vshift)}] (state) {
        \begin{tabular}{c}
          \footnotesize UAV State \\
          \footnotesize Estimator \\
    \end{tabular}};

    \node [block, below of=nmpc, shift = {(0.0cm, -\vshift)}] (rvc) {
        \begin{tabular}{c}
          \footnotesize Reciprocal Velocity \\
          \footnotesize Constraint Generator \\
    \end{tabular}};

    \node [block_single, below of=fcu, shift = {(0.0cm, -1.613*\vshift)}] (comm) {
        \begin{tabular}{c}
          \footnotesize Communication Module \\
    \end{tabular}};

    \node [block_single, above of=comm, shift = {(0.0cm, -0.012cm)}] (est) {
        \begin{tabular}{c}
          \footnotesize Estimation Module \\
    \end{tabular}};

  \end{pgfonlayer}



  \draw [arrow, ultra thick, dashed] ($(pmm.west) + (-1.5, -0.0)$) -- ($(pmm.west) + (0.0, 0.0)$) node[midway,above, shift={(-0.2, 0)}] {\footnotesize goal};
  \draw [arrow, ultra thick, dashed] ($(state.west) + (-1.5, -0.0)$) -- ($(state.west) + (0.0, 0.0)$) node[midway, below, shift={(-0.2, 0)}] {\begin{tabular}{c}
          \footnotesize sensor \\
          \footnotesize data \\
    \end{tabular}};
  \draw [arrow, ultra thick] ($(state.north) + (-0.0, -0.0)$) -- ($(pmm.south) + (0.0, 0.0)$) node[midway,left, shift={(-0.0, 0)}] {\footnotesize $\bm{p}_i, \bm{v}_i$};
  \draw [arrow, ultra thick] ($(rvc.north) + (-0.0, -0.0)$) -- ($(nmpc.south) + (0.0, 0.0)$) node[midway, left, shift={(-0.0, 0)}] {\footnotesize $A_{m,k}, b_{m,k}$};
  \draw [arrow, ultra thick] ($(pmm.east) + (-0.0, -0.0)$) -- ($(nmpc.west) + (0.0, 0.0)$) node[midway,above, shift={(-0.0, 0)}] {\footnotesize $\bm{T}_{ref, pmm}$};
  \draw [arrow, ultra thick] ($(nmpc.east) + (-0.0, -0.0)$) -- ($(fcu.west) + (0.0, 0.0)$) node[midway, above, shift={(-0.0, 0)}] {\footnotesize $\angularvel, \bodythrustvector$};
  \draw [arrow, ultra thick] ($(state.east) + (-0.0, -0.0)$) -- ($(rvc.west) + (0.0, 0.0)$) node[midway,above, shift={(-0.0, 0)}] {\footnotesize $\bm{p}_i, \bm{v}_i$};
  \draw [arrow, ultra thick] ($(rvc.east) + (1.0, 0.0)$) -- ($(rvc.east) + (0.0, 0.0)$) node[midway,above, shift={(0.1, 0)}] {\footnotesize $\bm{p}_m, \bm{v}_m$};
  \draw [arrow, ultra thick] ($(fcu.east) + (-0.0, -0.0)$) -- ($(actuators.west) + (0.0, 0.0)$) node[midway,above, shift={(-0.0, 0)}] {\footnotesize};
  \draw [arrow, ultra thick, dashed] ($(comm.east) + (2.2, -0.0)$) -- ($(comm.east) + (0.0, 0.0)$) node[midway, above, shift={(0.2, -0.10)}] {\begin{tabular}{c}
          \footnotesize messages from \\
          \footnotesize other UAVs \\
    \end{tabular}};

  \draw [arrow, ultra thick, dashed] ($(est.east) + (2.2, -0.0)$) -- ($(est.east) + (0.0, 0.0)$) node[midway, above, shift={(0.2, -0.10)}] {\begin{tabular}{c}
          \footnotesize sensor \\
          \footnotesize data \\
    \end{tabular}};

  \draw [ultra thick] ($(est.west) + (-0.0, -0.0)$) -- ($(est.west) + (-0.5, 0.0)$) node[midway,above, shift={(-0.0, 0)}] {\footnotesize};
  \draw [ultra thick] ($(comm.west) + (-0.0, -0.0)$) -- ($(comm.west) + (-0.5, 0.0)$) node[midway,above, shift={(-0.0, 0)}] {\footnotesize};

  \fill (7.2,-1.8) circle[radius=3pt];
  \draw[fill=white, draw=black] (7.8,-1.308) circle[radius=3pt];
  \draw[fill=white, draw=black] (7.8,-2.292) circle[radius=3pt];
  \draw [ultra thick] (7.2, -1.8) -- (7.8, -2.292) node[midway,above, shift={(-0.0, 0)}] {\footnotesize};
  \draw[fill=white, draw=black] (7.8,-1.308) circle[radius=3pt];
  \draw[fill=white, draw=black] (7.8,-2.292) circle[radius=3pt];
  \draw[<->] (7.55, -2.00) to[out=60, in=-60] node[above] {} (7.55,-1.6);


   \begin{pgfonlayer}{background}
     \path (pmm.west |- pmm.north)+(-0.35,0.35) node (a) {};
     \path (rvc.south -| rvc.east)+(+0.35,-0.35) node (b) {};
     \path[fill=color_grey!35,rounded corners, draw=none, densely dotted]
     (a) rectangle (b);
   \end{pgfonlayer}

    \node [below of=rvc, shift={(1.3,0.35)}] (freq) { \footnotesize \textbf{$\mathbf{\approx100}$ Hz}};


\end{tikzpicture}
  \vspace*{-0.1cm}
  \caption{Block diagram representing a single robot control and navigation pipeline for robot $i$ including the proposed approach for mutual collision avoidance for agile UAV flight.}
  \vspace*{-0.2cm}
  \label{fig:pipeline}
\end{figure*}

In addition to the vehicle dynamics, the proposed approach relies on established concepts of velocity obstacles and optimal reciprocal collision avoidance.
The concept of velocity obstacles, first introduced in~\cite{fiorini1998velocityobstacles}, was developed for motion planning in environments with dynamic obstacles and was further adapted for multi-robot scenarios by introducing \acp{rvo}~\cite{vandenberg2008rvo} accounting for scenarios with velocity obstacles induced by intelligent robots making decisions based on perceived environment.     
The velocity obstacle $VO_{A|B}^\tau$ for robot $A$ induced by obstacle $B$ for time window~$\tau$ is defined as a set of relative velocities of $A$ with respect to $B$ that will result in a collision during time window $\tau$.
Formally $VO_{A|B}^\tau$ can be described as
\begin{equation}\label{eq:velocity_obstacle}
  VO_{A|B}^\tau = \{\bm{v} | \exists t \in [0, \tau], ||\bm{p}_A + \bm{v}t - \bm{p}_B|| < r_A + r_B\},   
\end{equation}
where $p_A, p_B$, $r_A, r_B$ are positions and radii of robot $A$ and obstacle $B$, respectively. 
Hence, if $\bm{v}_A - \bm{v}_b \notin VO_{A|B}^\tau$, robot $A$ is guaranteed not to collide with $B$ in time window $[0, \tau]$.

The velocity obstacle can be extended to the concept of optimal reciprocal collision avoidance~\cite{vandenberg2010orca} by considering both $A$ and $B$ active decision-making agents, as follows.
The set of collision-avoiding velocities for robot $A$ given that robot $B$ selects a velocity from set $V_B$ is defined as 
\begin{equation}
  CA_{A|B}^\tau(V_B) = \{ \bm{v} | \bm{v} \notin VO_{A|B}^\tau \oplus V_B\},
\end{equation}
where operator $\oplus$ denotes Minkovski sum. 
The problem of finding sets of velocities for optimal reciprocal collision avoidance $ORCA_{A|B}^\tau$ and $ORCA_{B|A}^\tau$ is described as finding sets of permitted velocities $V_A^*$, $V_B^*$ that fulfill the following conditions: (i) the sets are reciprocal collision avoiding, thus 
   \begin{equation}
     CA_{A|B}^\tau(V_B^*) = V_A^* \text{ and } CA_{B|A}^\tau(V_A^*) = V_B^*,
   \end{equation}
   (ii) the sets maximize the intersection with velocities close to target velocities $\bm{v}_A^t, \bm{v}_B^t$. With $r_A = r_B = r$, this reads  
   \begin{equation}
     \begin{split}
       |ORCA_{A|B}^\tau \cap D(\bm{v}_A^t, r)| = |ORCA_{B|A}^\tau \cap D(\bm{v}_B^t, r)| \\ 
       \geq \min (V_A \cap D(\bm{v}_A^t, r), V_B \cap D(\bm{v}_B^t, r)) \\ 
       \forall\,\, V_A \subset CA_{A|B}^\tau(V_B), V_B \subset CA_{B|A}^\tau(V_A), r > 0\text{,}
     \end{split}
   \end{equation}
    where 
    \begin{equation}
      D(\bm{x}, y) = \{\bm{z} |\,\, y \geq |\bm{z} - \bm{x}|\}.
    \end{equation}
The set of velocities fulfilling these conditions can be constructed as
\begin{equation}\label{eq:orca_half}
  ORCA_{A|B}^\tau = \left\{\bm{v}\, \left| \left(\bm{v} - \bm{v}_A^t - \frac{1}{2}\bm{u}\right)\cdot \bm{n} \geq 0\right.\right\}\text{,}
\end{equation}
with $\bm{u}$ being the smallest change to relative velocity to avoid a collision on horizon $\tau$, specified as
\begin{equation} \label{eq:orca_u}
  \bm{u} = \left(\text{argmin}_{\bm{v} \in \partial VO_{A|B}^\tau} ||\bm{v} - (\bm{v}_A^t - \bm{v}_B^t)||\right) - (\bm{v}_A^t - \bm{v}_B^t),
\end{equation}
and $\bm{n}$ being the outward normal of $\partial VO_{A|B}^\tau$ at point $ (\bm{v}_A^t - \bm{v}_B^t) + \bm{u}$.  
$ORCA_{A|B}^\tau$ as specified in \eqref{eq:orca_half} ensures that robots $A$ and $B$ both contribute to avoiding mutual collisions in an equal way.  
For more details on velocity obstacles and optimal reciprocal collision avoidance, we refer to detailed descriptions provided in~\cite{fiorini1998velocityobstacles, vandenberg2008rvo,vandenberg2010orca}.



\section{Methodology}
\label{sec:methodology}

The proposed approach for high-speed mutual collision avoidance consists of several modules that process the requested goal destination, sensor data, and eventually, telemetries of other robots to generate quadrotor control inputs that result in collision-free navigation to the goal destination in multi-robot scenarios.
The block diagram of the pipeline is provided in \autoref{fig:pipeline}.

The necessary inputs of the designed pipeline include sensor data, which are processed by \textit{UAV State Estimator}, providing the estimates of the current robot's position and velocity.  
This estimate is supplied together with a user-provided goal destination to the \textit{PMM Reference Trajectory Generator} \cite{teissing2024pmm}, which computes minimum-time trajectory leading from the current state to a goal destination while respecting given kinematic constraints and ignoring collisions of any type. 
Simultaneously, the estimate of the current state of the robot, along with positions and velocities of other robots, is provided to the \textit{Reciprocal Velocity Constraint Generator}, which generates a set of linear reciprocal velocity constraints ensuring mutual collision avoidance among robots.
The positions and velocities of other robots are obtained either through \textit{Communication Module} or estimated from the robot's sensor data using an \textit{Estimation Module}.  
The generated reciprocal velocity constraints and the reference trajectory serve as inputs to \textit{NMPC Controller} generating control inputs that are passed to the \textit{Flight Control Unit}, which translates this reference to control commands for individual rotors.      
A detailed description of individual modules is provided in the following sections.

\subsection{Reference trajectory generation}

The reference trajectory $T_{ref}$ for \ac{nmpc} controller is generated using a Point-Mass Model (PMM) minimum-time trajectory generation approach introduced in~\cite{teissing2024pmm} \rev{chosen to balance the focus on minimization of flight time, feasibility of the provided trajectory considering kinematic constraints, and low computational demands}.  
The approach is used for generation of trajectories starting from initial conditions given by the estimated robot's position $\bm{p}_i$ and velocity $\linearvelvector_i$ to a final configuration given by the goal destination $\bm{g}_i$ and velocity $\bm{v}_{g,i}$.  
The feasibility of the produced control reference up to the second derivative of position is achieved through applying kinematic constraints given by limits on the norm of applied acceleration $\bm{a}$ and velocity $\bm{v}$. 

The generated trajectory is represented as $T_{ref,pmm} = \left(t_{p,0}, \dots t_{p,K}\right)$, where every transition point is represented by tuple $t_{p,k} = \{\bm{p}_p, \bm{v}_p, \bm{a}_p\}$ encoding reference position, velocity and acceleration of particular transition point. 
For generation of the full state reference for NMPC control problem formulation, the reference trajectory $T_{ref,pmm}$ is augmented with rotational part of the quadrotor's state~\cite{gupta2025lolnmpclowleveldynamicsintegration} yielding the final reference trajectory $T_{ref} = \left(t_{0}, \dots t_{K}\right)$ where $t_k = \{\bm{p}_p, \bm{q}, \bm{v}_p, \bm{\omega}\}$ with quaternion $\bm{q}$ representing quadrotor's orientation, and $\bm{\omega}$ representing angular velocities. 

\subsection{NMPC controller with reciprocal velocity constraints}

The introduced control problem is formulated under the \ac{nmpc} framework as follows:
	\begin{align}
    & \hspace*{-0.9cm} \minimize_{\substack{\bm{u}_{0} \ldots \bm{u}_{N-1}}} \sum_{k=1}^{N} ||\Delta \bm{x}_{k}||^{2}_{Q} + ||\Delta \bm{u}_{k-1}||^{2}_{R} + ||\bm{s}_{k}||^{2}_{Z},\label{eq:nmpc_cost}\\
    \text{s.t. } & \bm{x}_{0} = \bm{x}(0),\label{eq:nmpc_init_c}\\
    \bm{x}_{k+1} & = f_{dyn}(\bm{x}_{k},\bm{u}_{k}) ,\,\, k \in \{0, \dots, N-1\},\label{eq:nmpc_dyn_c}\\
                 \angularvel_{min} & \leq \angularvel_k \leq \angularvel_{max},\,\, k \in \{1, \dots, N\},\label{eq:nmpc_angular_c}\\
                 \motorforce_{min} & \leq \motorforce_{i,k} \leq \motorforce_{max},\, i \in \{1, 2, 3, 4\},\, k \in \{0, \dots, N-1\},\label{eq:nmpc_thrust_c}\\
                F_{min} & \leq \sum_{i=1}^4 \motorforce_{i,k} \leq F_{max},\, k \in \{0, \dots, N-1\},\label{eq:nmpc_collective_thrust_c}\\
                 b_{m,k} & \leq A_{m,k} \linearvelvector_k + s_{m,k} ,\, m \in \{1, \dots, M\},\, k \in \{1, \dots, N\},\label{eq:nmpc_vel_c}
	\end{align}
where $N$ is the number of transition points, $M$ is the number of robots considered for mutual collision avoidance, $\Delta \bm{x}_k = \bm{x}_k - \bm{x}_{ref, k}$, $\Delta \bm{u}_k = \bm{u}_k - \bm{u}_{ref}$, \rev{and} $\bm{s}_k = \begin{bmatrix} s_{1,k},\dots, s_{M,k}\end{bmatrix}$ \rev{is a vector of slack variables}. The $\bm{x}_{ref, k}$ stands for a reference state at $k$-th transition point given by an equivalent segment of reference trajectory $T_{ref}$, and $\bm{u}_{ref}$ represents the reference input motor forces.   
$Q \succeq 0, R \succeq 0, Z \succeq 0$ stands for the state, input and slack variables weighting matrices, respectively, and expression $||\bm{y}||_W^2 = \bm{y}^TW\bm{y}$.
  The set of constraints consists of constraints on quadrotor's initial state~\eqref{eq:nmpc_init_c}, dynamic model constraints~\eqref{eq:nmpc_dyn_c} where $f_{dyn}(\cdot)$ corresponds to dynamic model discretized using the Runge-Kutta method, constraints on angular rates $\angularvel$~\eqref{eq:nmpc_angular_c}, limits on individual motor thrusts $\motorforce_i$~\eqref{eq:nmpc_thrust_c}, limits on collective motor thrust~\eqref{eq:nmpc_collective_thrust_c}, and time-dependent \rev{soft} linear velocity constraints for mutual collision avoidance~\eqref{eq:nmpc_vel_c} described in the following section. 

    Unlike \ac{mpc}-based approaches applying \ac{orca}~\cite{cheng2017DecentralizedNavigationMultiple, arul2020DCADDecentralizedCollision}, the proposed approach does not compute velocity constraints for every transition point on the prediction horizon, thereby reducing computational demands without degrading performance. 
Given only limited information from other UAVs (current position and velocity) and the high agility of motion, the most reliable constraints are those derived from the current state.
The feasible collision-free velocity set defined by the reciprocal velocity constraints is always convex. 
    Therefore, any convex combination of feasible velocities also remains feasible. 
    If a robot applies a sequence of feasible velocities over a time interval $t_x$, the resulting displacement is equivalent to applying a constant velocity equal to the average velocity over that interval. 
    Due to the convexity of the feasible set, this average velocity is also feasible. 
Consequently, applying a single set of reciprocal velocity constraints over the entire prediction horizon, while allowing the velocity to vary between individual transition points, preserves the mutual collision-avoidance guarantees without requiring the constraints to be recomputed for every transition point.

  \subsection{Time-dependent reciprocal velocity constraints}
  The mutual collision avoidance is introduced in the \ac{nmpc} controller through reciprocal collision avoidance constraints~\eqref{eq:nmpc_vel_c} generated as follows.
    Given the current position $\bm{p}_i$ and velocity $\bm{v}_i$ of the robot with index $i$, the set of velocities for optimal collision avoidance $ORCA_{i|j}^\tau$ is computed for every neighboring robot $j \in \{1, \dots, M\}$, with position $\bm{p}_j$, collision radius $r_{ca}$, and velocity $\bm{v}_j$, where current velocities $\bm{v}_i$, and $\bm{v}_j$ are considered as target velocities in computation of $ORCA_{i|j}^\tau$.
  The individual sets of velocities $ORCA_{i|j}^\tau$ are then converted to linear constraints of the form  
\begin{equation} \label{eq:orca_lin_constraints}
  b_m \leq A_m \bm{v}, 
\end{equation}
with 
\begin{equation}\label{eq:orca_lin_constraints_am}
  A_m = \frac{\bm{u}_{m}}{|\bm{u}_{m}|},~~\,\, b_m =  \frac{\bm{u}_{m}}{|\bm{u}_{m}|} \cdot \left(\bm{v}_i + \frac{\bm{u_m}}{2}\right),
\end{equation}
where $\bm{u}_{m}$ is computed in compliance with optimal reciprocal collision avoidance concept~\eqref{eq:orca_u} as  
\begin{equation}
  \bm{u}_{m} = \left(\text{argmin}_{\bm{v} \in \partial VO_{i|j}^\tau} ||\bm{v} - (\bm{v}_i - \bm{v}_j)||\right) - (\bm{v}_i - \bm{v}_j)
\end{equation}
with $VO_{i|j}^\tau$ defined according to~\eqref{eq:velocity_obstacle}.
To cope with the latency of data resulting from communication delays and lower frequencies of incoming messages compared to the control loop frequency, the first-order linear motion model is applied to predict current positions of other \acp{uav} based on the most recent available information.   

\begin{figure}[htb]
  \centering
  \input{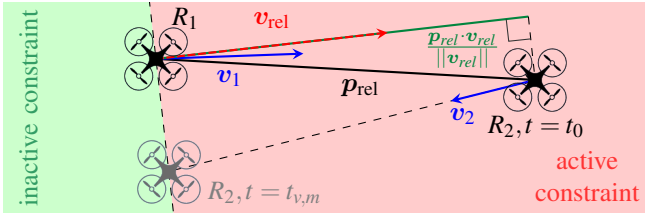}
  \caption{The illustration of the introduced time validity of the reciprocal velocity constraints.}
  \vspace*{-0.3cm}
  \label{fig:time_validity}
\end{figure}

    Due to the absence of information about future trajectories of other robots, the constraint~\eqref{eq:orca_lin_constraints} represents the only velocity constraint that can be computed based on the available information --- current position and velocity of the robots $i$ and $j$.
While applying this constraint to the entire control horizon provides the required mutual collision avoidance guarantees (given $\tau \geq T_h$), such an approach is unnecessarily restrictive and hinders the performance of the method.   
Therefore, we introduce the time validity $t_{v,m}$ of the velocity constraint for robot $m$ whose estimate is given by 
\begin{equation}\label{eq:time_validity}
  t_{v,m} = \max\left(\frac{\bm{p}_{rel} \cdot \bm{v}_{rel}}{||\bm{v}_{rel}||^2}, 0\right),
\end{equation}
    which represents the time after which the angle between vector \rev{$\bm{p}_{rel} = \bm{p}_j - \bm{p}_i$}, representing the relative position between robots, and vector \rev{$\bm{v}_{rel} = \bm{v}_j - \bm{v}_i$}, representing robots' relative velocity, exceeds $\frac{\pi}{2}$, i.e., the constrain\rev{t} is applied only for time steps before the drone \rev{would pass} the possibly colliding drone \rev{assuming both robots maintain their current velocities $\bm{v}_i, \bm{v}_j$} (see~\autoref{fig:time_validity}).
    Given the time validity~\eqref{eq:time_validity} for each constraint, we introduce \acp{rvc} for mutual collision avoidance in \ac{nmpc} formulation as time dependent variable constraints given by
\begin{equation}\label{eq:vel_constraints_time_dependent}
  \begin{split}
    A_{m,k} &= \begin{cases}
      A_m & \text{if}\ \ \ t_k \leq t_{v,m}, \\
      \left[0\right]_{3x3} & \text{if}\ \ \ t_k > t_{v,m},
    \end{cases}\\ 
    b_{m,k} &= \begin{cases}
      b_m & \text{if}\ \ \ t_k \leq t_{v,m}, \\
      0 & \text{if}\ \ \ t_k > t_{v,m},
    \end{cases}
  \end{split}
\end{equation}
where $t_k$ is the time corresponding to $k$-th transition point in the control horizon.

    The introduced time-dependent reciprocal collision-avoidance constraints~\eqref{eq:vel_constraints_time_dependent} are applied as soft constraints with slack variables $s_{m,k}$ in the presented \ac{nmpc} formulation~\eqref{eq:nmpc_cost}. This prevents infeasibility of the problem and thus achieves practicality of the approach in real-world scenarios with uncertainties and an arbitrary number of robots. 

\section{Results}
\label{sec:results}
This section presents statistical analysis, ablation studies, and experimental results  
demonstrating the performance of the proposed approach.
In all presented results, the time of the transition of a robot $r_i$ from point $\mathbf{a}$ to point $\mathbf{b}$ is specified as a difference between time $t_{a,i}$ when the request to navigate to point $\mathbf{b}$ was received and time when the $\mathbf{b}$ is considered to be reached, which is defined as 
\begin{equation}
  t_{b, i} = \min \left\{t | \forall t_x > t,\,\, ||\positionvector(t_x) - \mathbf{b}|| \leq \epsilon\right\}.
\end{equation}
The time of the transition of a set of robots with indices $i \in \mathbb{I}_R$ from initial to goal configurations is defined as
\begin{equation}
  T_x = \max_{i \in \mathbb{I}_R} t_{b,i} - \min_{i \in \mathbb{I}_R} t_{a,i}.
\end{equation}
Unless otherwise specified, evaluations were performed in a simulation on a computer with a 8-core AMD Ryzen 7 5800X CPU with base frequency \SI{3.80}{GHz}, $\epsilon = \SI{0.1}{\meter}$, $\tau=\SI{8}{\second}$, collision radius $r_{ca} = \SI{2}{\meter}$, asynchronous communication, and robots considered to be spheres with radius \SI{0.25}{\meter} for collision avoidance evaluation.

\subsection{Performance test}\label{sec:performance_test}
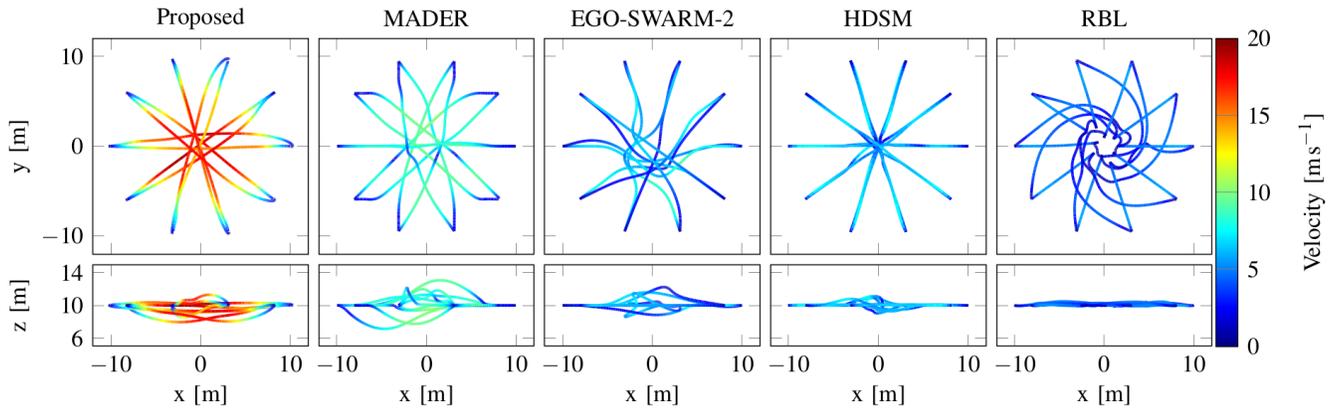
\begin{figure*}[htb]
  \centering


\begin{tikzpicture}

  \pgfplotstableread[col sep=comma]{./figs/tikz_data/sample_trajectories_our_approach.csv}{\tablemine}
  \pgfplotstableread[col sep=comma]{./figs/tikz_data/sample_trajectories_mader.csv}{\tablemader}
  \pgfplotstableread[col sep=comma]{./figs/tikz_data/sample_trajectories_egoswarm2.csv}{\tableego}
  \pgfplotstableread[col sep=comma]{./figs/tikz_data/sample_trajectories_hdsm.csv}{\tablehdsm}
  \pgfplotstableread[col sep=comma]{./figs/tikz_data/sample_trajectories_rbl.csv}{\tablerbl}


  \pgfmathsetmacro{\zheight}{0.15}

  \begin{groupplot}[
    group style={
        group size=5 by 2,
        horizontal sep=2cm,
        x descriptions at=edge bottom,
        y descriptions at=edge left,
        vertical sep=4pt,
        horizontal sep=4pt,
    },
    point meta min=0.0,                         
    point meta max=20.0,                         
    width=0.25\textwidth,
    height=0.25\textwidth,
    xlabel={x [m]},
    ylabel={y [m]},
    xmin=-12, xmax=12,
    ymin=-12, ymax=12,
    colormap/jet,
    /pgfplots/table/ignore chars={|},
    title style={yshift=-5pt}, 
    x label style={yshift=0.8ex},
    y label style={yshift=-0.8ex},
    label style={font=\small},
    tick label style={font=\small},
    legend style={font=\small},
    title style={font=\small}
]

\def\plotTrajectories#1{
    \foreach \i in {1,...,10} {
        \addplot[mesh, point meta=explicit, line width =1.0, shader=interp] table [
            x=x\i, y=y\i, meta=v\i
        ] {#1};
    }
}

\def\plotTrajectoriesZ#1{
    \foreach \i in {1,...,10} {
        \addplot[mesh, point meta=explicit, line width =1.0, shader=interp] table [
            x=x\i, y=z\i, meta=v\i
        ] {#1};
    }
}

\nextgroupplot[
    title={Proposed},
    title style={yshift=-1pt}, 
    y label style={yshift=-0.8ex},
]
\plotTrajectories{\tablemine};

\nextgroupplot[
    title={MADER},
]
\plotTrajectories{\tablemader};

\nextgroupplot[
    title={EGO-SWARM-2},
]
\plotTrajectories{\tableego};

\nextgroupplot[
    title={HDSM},
]
\plotTrajectories{\tablehdsm};

\nextgroupplot[
    title={RBL},
    colorbar,
    point meta min=0.0,                         
    point meta max=20.0,                         
    colorbar style={
        at={(1.02,0.286)},
        anchor=west,
        ylabel={Velocity [\si{\meter\per\second}]},
        width=0.1*\pgfkeysvalueof{/pgfplots/parent axis width},
        height=1.429*\pgfkeysvalueof{/pgfplots/parent axis height},
        ylabel style={shift={(0.0, -0.58)}},
    },
]
\plotTrajectories{\tablerbl};

\nextgroupplot[
    ylabel={z [m]},
    ymin=5, ymax=15,
    height=\zheight\textwidth,
    ytick={6.00, 10.0, 14.0},
    y label style={yshift=-0.8ex},
]
\plotTrajectoriesZ{\tablemine};

\nextgroupplot[
    height=\zheight\textwidth,
    ymin=5, ymax=15,
    ytick={6.00, 10.0, 14.0},
]
\plotTrajectoriesZ{\tablemader};

\nextgroupplot[
    height=\zheight\textwidth,
    ymin=5, ymax=15,
    ytick={6.00, 10.0, 14.0},
]
\plotTrajectoriesZ{\tableego};

\nextgroupplot[
    height=\zheight\textwidth,
    ymin=5, ymax=15,
    ytick={6.00, 10.0, 14.0},
]
\plotTrajectoriesZ{\tablehdsm};

\nextgroupplot[
    height=\zheight\textwidth,
    ymin=5, ymax=15,
    ytick={6.00, 10.0, 14.0},
]
\plotTrajectoriesZ{\tablerbl};

\end{groupplot}

\end{tikzpicture}

  \caption{Qualitative comparison of trajectories generated by individual approaches in scenario involving 10 UAVs navigating to antipodal positions on the circle of radius \SI{10}{\meter} with velocity and acceleration limits \SI{20}{\meter\per\second} and \SI{40}{\meter\per\second\squared}, respectively.}
  \vspace*{-0.3cm}
  \label{fig:trajectories_comparison}
\end{figure*}

We demonstrate the performance of the proposed approach in a challenging scenario involving 10 \acp{uav} simultaneously navigating to antipodal positions on a circle of radius \SI{10}{\meter} (further referred as APCX scenario) with $r_{ca} = \SI{0.6}{\meter}$, and velocity and acceleration constraints up to \SI{20}{\meter\per\second} and \SI{40}{\meter\per\second\squared}, respectively. These constraints do not stem from the limitations in the proposed algorithm but reflect the constraints of the platform used for real-world evaluation.
The obtained results show that the proposed approach outperforms other state-of-the-art approaches, MADER~\cite{tordesillas2022MADERTrajectoryPlannera}, EGO-SWARM-2~\cite{swarmingInTheWild}, HDSM~\cite{toumieh2024motionPlanning}, and RBL~\cite{boldrer2024rulebasedlloydalgorithmmultirobot} in terms of flight time while achieving collision-free navigation in all scenarios.
Our approach shows \SI{31}{\percent} reduction compared to the best-performing approach, HDSM, even though HDSM does not consider quadrotor's dynamics.
In the comparison, the proposed approach and RBL~\cite{boldrer2024rulebasedlloydalgorithmmultirobot} were deployed in a simulator that simulates the full dynamics of \acp{uav} and uncertainties in state estimation, whereas the rest of the methods were evaluated in authors' environments mostly assuming perfect control.   
The detailed results are presented in~\autoref{tab:comparison} with qualitative comparison provided in \autoref{fig:trajectories_comparison}. 

Since MADER and HDSM apply kinematic constraints per axis, the comparison includes also setups in which the original values of kinematic constraints are applied per axis instead of the limit to their norms.
This allows these approaches to mitigate their disadvantage of being overly restricted in certain directions of flight, while effectively enabling violation of the kinematic constraints by up to factor $\sqrt{3}$ depending on direction of flight.  
Even under this extremely unfair setup, our approach reduces flight time by more than $\SI{11}{\percent}$.  

\begin{remark}
  Due to the different ways of applying kinematic constraints, solving the problem in a different domain (control vs. planning), simplifying dynamic models, and assuming perfect control applied by individual methods, the comparison cannot be entirely fair. However, the majority of unfairness is in favor of other approaches rather than our \rev{approach considering full quadrotor's dynamic model, and uncertainties in both state estimation and control}.
\end{remark}

\begin{table*}[t]
\renewcommand{\arraystretch}{0.90}
\setlength\tabcolsep{4.5pt}
  \begin{center}
    \caption{Comparison of approaches for the solution of a simultaneous navigation of 10 \acp{uav} to antipodal positions on a circle of radius \SI{10}{\meter}. The results are averaged over 100 trials. The robots are considered to be spheres with a radius \SI{0.25}{\meter} for collision avoidance evaluation. The individual approaches have been parametrized with a focus on minimization of flight time while keeping the success rate close to $100\%$. The experienced collisions, deadlocks and optimization failures causing drops in success rates could be avoided by choosing another parametrization as the authors of all compared methods demonstrated $100\%$ success rate in evaluations of their methods. 
The terms \textit{norm} and \textit{per axis} indicate whether constraints are scaled by $1/\sqrt{3}$ to bound the norm, or applied directly per axis (yielding an effective norm limit of $\sqrt{3}$ times the given value).
    }\label{tab:comparison}
    \newcommand*{\hshift}{\hspace*{0.0cm}}
    \begin{tabular}{l c c c c c c c c c c}
      \toprule 
      \multirow{2.4}{*}{Approach} & max. vel. & \multirow{2.4}{*}{\begin{tabular}{cc} Success \\ rate [\%]\end{tabular}} & \multicolumn{2}{c}{Flight time [\si{\second}]} &  \multicolumn{2}{c}{Flight distance [\si{\meter}]} & \multicolumn{2}{c}{Flight velocity [\si{\meter\per\second}]} & \multicolumn{2}{c}{Min. mutual dist. [\si{\meter}]} \\
      \cmidrule(lr){2-2} \cmidrule(lr){4-5} \cmidrule(lr){6-7} \cmidrule(lr){8-9} \cmidrule(lr){10-11}
      & max. acc. & & ~~mean & ~std. dev. & ~~mean & ~std.dev & ~~~~mean & std.dev & ~~~mean & min\\ 
      \midrule
      RBL~\cite{boldrer2024rulebasedlloydalgorithmmultirobot} & \multirow{8}{*}{\begin{tabular}{cc} \SI{10}{\meter\per\second} \\\midrule \SI{7}{\meter\per\second\squared} \end{tabular}} & \textbf{100.0} & ~~11.71 & ~0.92 & ~~26.28 & ~0.75 & ~~~~2.54 & 0.05 & ~~~0.86 & 0.61\\
      EGO-SWARM-2~\cite{swarmingInTheWild} & & 99.0 & ~~10.67 & ~0.87 & ~~\textbf{20.06} & ~0.11 & ~~~~2.42 & 0.13 & ~~~0.59 & 0.47\\
      MADER~\cite{tordesillas2022MADERTrajectoryPlannera} - norm & & 96.0 & ~~~7.89 & ~0.52 & ~~20.82 & ~0.18 & ~~~~3.14 & 0.09 & ~~~0.72 & 0.19\\
      MADER~\cite{tordesillas2022MADERTrajectoryPlannera} - per axis &  & 95.0 & ~~~6.14 & ~0.37 & ~~20.89 & ~0.26 & ~~~~4.04 & 0.12 & ~~~0.70 & 0.10\\
      HDSM~\cite{toumieh2024motionPlanning} - norm & & \textbf{100.0} & ~~~7.96 & ~0.18 & ~~20.35 & ~\textbf{0.04} & ~~~~2.87 & \textbf{0.03} & ~~~0.53 & 0.51\\
      HDSM~\cite{toumieh2024motionPlanning} - per axis & & \textbf{100.0} & ~~~5.20 & ~0.19 & ~~20.34 & ~0.09 & ~~~~4.40 & 0.09 & ~~~0.56 & 0.51\\
      Proposed & & \textbf{100.0} & ~~~\textbf{4.80} & ~\textbf{0.09} & ~~20.65 & ~0.05 & ~~~~\textbf{4.60} & \textbf{0.03} & ~~~1.12 & \textbf{1.04}\\
      Proposed with drag & & \textbf{100.0} & ~~~5.91 & ~0.12 & ~~22.78 & ~0.12 & ~~~~4.05 & 0.04 & ~~~\textbf{1.20} & 0.98\\
      \midrule
      RBL~\cite{boldrer2024rulebasedlloydalgorithmmultirobot} & \multirow{8}{*}{\begin{tabular}{cc} \SI{20}{\meter\per\second} \\\midrule \SI{40}{\meter\per\second\squared} \end{tabular}} & \textbf{100.0} & ~~11.59 & ~0.74 & ~~26.20 & ~0.83 & ~~~~2.54 & 0.05 & ~~~0.87 & 0.63\\
      EGO-SWARM-2~\cite{swarmingInTheWild} & & 99.0 & ~~~8.16 & ~0.50 & ~~21.15 & ~0.37 & ~~~~3.61 & 0.22 & ~~~0.67 & 0.47\\
      MADER~\cite{tordesillas2022MADERTrajectoryPlannera} - norm & & 98.0 & ~~~5.28 & ~0.35 & ~~20.93 & ~0.20 & ~~~~4.70 & 0.16 & ~~~0.72 & 0.51\\
      MADER~\cite{tordesillas2022MADERTrajectoryPlannera} - per axis & & 94.0 & ~~~5.31 & ~0.53 & ~~21.13 & ~0.21 & ~~~~4.75 & 0.17 & ~~~0.75 & 0.38\\
      HDSM~\cite{toumieh2024motionPlanning} - norm & & \textbf{100.0} & ~~~4.46 & ~0.19 & ~~20.50 & ~0.10 & ~~~~4.78 & 0.18 & ~~~0.67 & 0.56\\
      HDSM~\cite{toumieh2024motionPlanning} - per axis & & \textbf{100.0} & ~~~3.43 & ~0.17 & ~~\textbf{20.43} & ~0.10 & ~~~~6.96 & 0.16 & ~~~0.67 & 0.57\\
      Proposed & & \textbf{100.0} & ~~~\textbf{3.07} & ~0.08 & ~~21.15 & ~0.06 & ~~~~\textbf{7.36} & 0.06 & ~~~\textbf{0.98} & 0.81\\
      Proposed with drag & & \textbf{100.0} & ~~~3.84 & ~\textbf{0.06} & ~~21.62 & ~\textbf{0.05} & ~~~~5.93 & \textbf{0.05} & ~~~\textbf{0.98} & \textbf{0.82}\\
      \bottomrule
    \end{tabular}
  \end{center}
  \vspace*{-0.4cm}
\end{table*}

\subsection{Robustness to communication latency, imprecisions in state estimation and communication dropouts} 

The robustness of the proposed approach to latency and noise in obtained states of other \acp{uav} is demonstrated through a series of simulations with modelled latency, decrease in frequency of incoming messages, and noise in estimated states of other \acp{uav}. 
While the evaluation is performed in simulation, the modelled errors emulate real-world conditions resulting from the usage of wireless means of communication or estimation of the state of \acp{uav} using onboard sensors and processing.

\begin{figure}[htb]
  \centering
  \pgfplotsset{compat=1.16} 

\begin{tikzpicture}

\pgfplotstableread[col sep=comma]{./figs/tikz_data/robustness_test_100_flights.csv}{\table}

 \pgfplotsset{
    colormap={redgreen}{rgb255(0cm)=(255,0,0); rgb255(0.5cm)=(255,255,0); rgb255(1cm)=(0,255,0)}
}

\begin{axis}[
    colorbar,                                  
    xlabel={Frequency [Hz]},                          
    ylabel={Delay [ms]},                          
    point meta min=0.0,                         
    point meta max=1.0,                         
    enlargelimits=false,                      
    tick label style={font=\small},           
    xtick={1,2,3,4,5,6},                      
    ytick={1,2,3,4,5,6},                      
    xticklabels={2,5,10,25,50,100},          
    yticklabels={0, 25, 50, 100, 200, 400}, 
    xmin=0.5, xmax=6.5,
    ymin=0.5, ymax=6.5,
    width=7.1cm, height=3.9cm,                
    nodes near coords style={font=\small},    
    every node near coord/.append style={     
      black, text opacity=1, anchor=center
    },
    colorbar style={
        ymin=0.0, ymax=1.0,                   
        ylabel={Success rate [-]},
    },
]

\addplot[
    matrix plot*,                             
    point meta=explicit,                      
    opacity=0.0,
    nodes near coords,                        
] table [
    x=x,                               
    y=y,                               
    meta=minMutualDistMean             
] {\table};

\addplot[
    matrix plot*,                             
    point meta=explicit,                      
] table [
    x=x,                               
    y=y,                               
    meta=successRate             
] {\table};

\end{axis}

\end{tikzpicture}
  \vspace*{-0.6cm}
  \caption{The success rate (shown in colors of the matrix) and minimum mutual distance between \acp{uav} (shown as numbers in the matrix [m]) under varying delay and frequency of messages obtained from other robots. The results for every delay-frequency pair are based on 100 flights involving 4 \acp{uav} in APCX scenario.}
  \label{fig:robustness_test}
\end{figure}
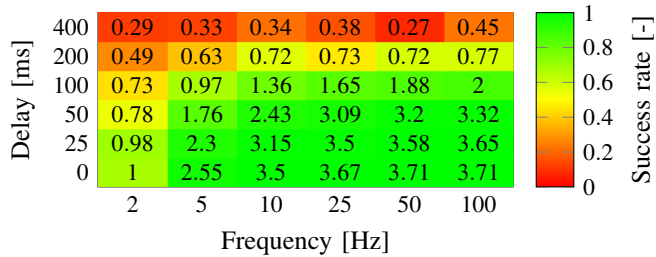

In the first set of simulations, the state of other \acp{uav} was delayed by up to \SI{400}{\milli\second} and its incoming frequency was decreased down to \SI{2}{\hertz} effectively causing additional dynamic latency of information about the current state of \acp{uav}.   
The results show that the approach is able to cope with delays of up to \SI{50}{\milli\second} and with frequencies down to \SI{10}{\hertz}, which is achievable both using standard means of wireless communication and \ac{uav} detection algorithms using onboard sensors.
The detailed results in~\autoref{fig:robustness_test} show a significant increase in the number of collisions for delays above \SI{100}{\milli\second} and frequencies below \SI{10}{\hertz}. 
Note that for such frequency-delay combinations and applied velocity and acceleration constraints, the \acp{uav} can move by more than \SI{4}{\meter} and change velocity by more than \SI{8}{\meter\per\second} by the time when the data are being processed.
While the algorithm can also be used with these delays, they must be considered in conjunction with the applied kinematic constraints when selecting the collision radius for velocity constraint generation.  

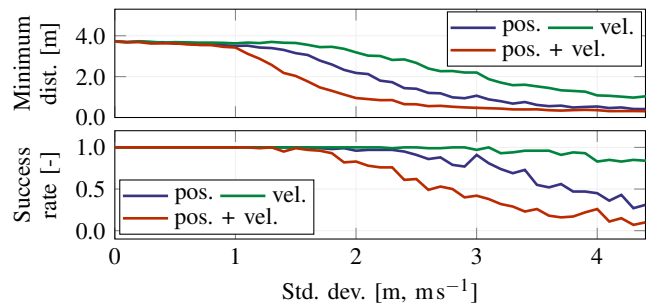
\begin{figure}[htb]
  \centering
  \definecolor{color_blue}{rgb}{0.22, 0.2, 0.502}
\definecolor{color_red}{rgb}{0.737,0.165,0}
\definecolor{color_green}{rgb}{0, .522, .243}
\hspace*{-0.22cm}
\begin{tikzpicture}[font=\normalsize]

  \pgfplotstableread[col sep=comma]{./figs/tikz_data/results_robustness_check_noise_pos.csv}{\tablepos}
  \pgfplotstableread[col sep=comma]{./figs/tikz_data/results_robustness_check_noise_vel.csv}{\tablevel}
  \pgfplotstableread[col sep=comma]{./figs/tikz_data/results_robustness_check_noise_posvel.csv}{\tableposvel}

  \begin{groupplot}[
    group style={
          group size=1 by 2,
          x descriptions at=edge bottom,
          vertical sep=5pt,
    },
    name=top,
    width=1.00\columnwidth,
    grid=major,
    grid style={draw=gray!12,line width=.1pt},
    xlabel= {Std. dev. [m, \si{\meter\per\second}]},
    xmin=0, xmax=4.4,
    y label style={yshift=-1.3ex},
    x label style={yshift=0.8ex},
    label style={font=\small},
    tick label style={font=\small},
    legend style={font=\small},
    ]

    \nextgroupplot[
      height=0.35\columnwidth,
      ylabel= {\begin{tabular}{c}Minimum \\ dist. [m] \end{tabular}},
      ymin=0, ymax=5.3,
      legend style={at={(0.9935,0.98)}, legend columns = 2, row sep=-0.7ex, inner xsep=0.7pt,
  inner ysep=0.2pt}, 
      yticklabels={,0.0, 2.0, 4.0},
    ]

    \addplot[color=color_blue, line width=1.0pt, opacity=1.0] table[y=minMutualDistMean, x=noiseStdPos] {\tablepos};
    \addplot[color=color_green, line width=1.0pt, opacity=1.0] table[y=minMutualDistMean, x=noiseStdVel] {\tablevel};
    \addplot[color=color_red, line width=1.0pt, opacity=1.0] table[y=minMutualDistMean, x=noiseStdVel] {\tableposvel};

    \addlegendentry{\small pos.}
    \addlegendentry{\small vel.}
    \addlegendentry{\makebox[0pt][l]{\hspace*{-0.26cm}\small pos. + vel.}}

  \nextgroupplot[
    ylabel= {\begin{tabular}{c}Success \\ rate [-] \end{tabular}},
    ymin=-0.1, ymax=1.2,
    height=0.35\columnwidth,
    legend style={at={(0.376,0.55)}, legend columns = 2, row sep=-0.7ex, inner xsep=0.7pt,
  inner ysep=0.2pt}, 
    yticklabels={,0.0, 0.5, 1.0},
  ]

  \addplot[name path=SP, color=color_blue, line width=1.0pt, opacity=1.0] table[y=successRate, x=noiseStdPos] {\tablepos};
  \addplot[name path=SV, color=color_green, line width=1.0pt, opacity=1.0] table[y=successRate, x=noiseStdVel] {\tablevel};
  \addplot[name path=SPV, color=color_red, line width=1.0pt, opacity=1.0] table[y=successRate, x=noiseStdVel] {\tableposvel};

  \addlegendentry{\small pos.}
  \addlegendentry{\small vel.}
  \addlegendentry{\makebox[0pt][l]{\hspace*{-0.26cm}\small pos. + vel.}}

  \end{groupplot}

\end{tikzpicture}
  \vspace*{-0.6cm}
  \caption{The minimum mutual distance between \acp{uav} and a success rate under varying noise in estimation of position and velocity of other robots. The results for every standard deviation are based on 100 flights involving 4 \acp{uav} in APCX scenario.}
  \label{fig:robustness_noise}
\end{figure}

In the second set of simulations, we analyze the influence of noise in estimates of other \acp{uav}' positions and velocities on the performance of the proposed method.
The results show that the method is capable of efficient operation in the presence of noise modelled as a Gaussian process $N(0,\,\sigma^{2})$ up to $\sigma_p = \SI{1}{\meter}$, $\sigma_v = \SI{2}{\meter\per\second}$ for position and velocity estimates, respectively.  
The detailed results are presented in \autoref{fig:robustness_noise}.

\subsection{Reliability test}
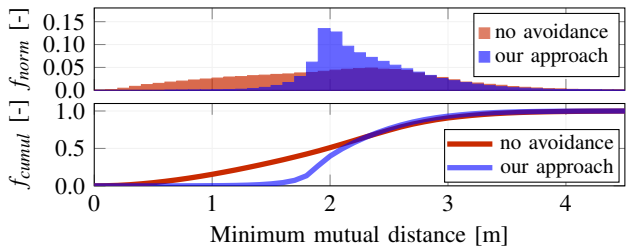
\begin{figure}[htb]
  \centering
  \begin{tikzpicture}[font=\normalsize]

  \definecolor{color_blue}{rgb}{0.22, 0.2, 0.802}
  \definecolor{color_red}{rgb}{0.787,0.165,0}
  \definecolor{color_green}{rgb}{0, .522, .243}

  \pgfplotstableread[col sep=comma]{./figs/tikz_data/rel_test_no_collision.csv}{\tablenoavoidance}
  \pgfplotstableread[col sep=comma]{./figs/tikz_data/rel_test_our_approach.csv}{\tableourapproach}

  \begin{groupplot}[
    group style={
          group size=1 by 2,
          x descriptions at=edge bottom,
          vertical sep=5pt,
    },
    name=top,
    width=1.0\columnwidth,
    height=0.5\columnwidth,
    grid=both,
    grid style={draw=gray!10,line width=.1pt},
    xlabel={Minimum mutual distance [m]},
    xmin=0, xmax=4.5,
    legend columns=1,
    legend style={cells={align=left}, text=black, row sep=-0.4ex, at={(0.998, 0.98)}},
    axis line style={latex-latex},
    scaled ticks=false,
    tick label style={/pgf/number format/fixed}, 
    label style={font=\small},
    tick label style={font=\small},
    legend style={font=\small},
    ]

    \nextgroupplot[
      height=0.310\columnwidth,
      ylabel={$f_{norm}$ [-]},
      y label style={yshift=-1.5ex},
      legend style={at={(0.990,0.94)}, inner sep=2pt, row sep=-2pt},
      ymin=0, ymax=0.18,
      yticklabels={,0.0, 0.05, 0.10, 0.15},
    ]

    \addplot+[forget plot, no markers, ybar interval, fill, color=color_red, fill opacity=0.6, opacity=0.7] table [x=Value, y=Frequency] {\tablenoavoidance};
    \addplot+[forget plot, no markers, ybar interval, fill, color=blue, fill opacity=0.4, opacity=0.6] table [x=Value, y=Frequency] {\tableourapproach};

  \addlegendimage{only marks, mark=square*, color=color_red, fill opacity=0.4, opacity=0.6}
  \addlegendimage{only marks, mark=square*, color=blue, fill opacity=0.4, opacity=0.6}
  \addlegendentry{\footnotesize no avoidance}
  \addlegendentry{\footnotesize our approach}

    \nextgroupplot[
      height=0.310\columnwidth,
      ylabel={$f_{cumul}$ [-]},
      y label style={yshift=-1.5ex},
      x label style={yshift=0.8ex},
      legend style={at={(0.990,0.69)}, inner sep=0pt, row sep=-2pt},
      ymin=0, ymax=1.10,
      yticklabels={,0.0, 0.5, 1.0},
    ]

    \addplot[no markers, color=color_red, line width=2.0, opacity=1.0] table [x=Value, y=Cumulative] {\tablenoavoidance};
    \addplot[no markers, color=blue, line width=2.0, opacity=0.6] table [x=Value, y=Cumulative] {\tableourapproach};

  \addlegendentry{\footnotesize no avoidance}
  \addlegendentry{\footnotesize our approach}


  \end{groupplot}

\end{tikzpicture}
\vspace{-0.1cm}
  \caption{Comparison of minimum mutual distances among \acp{uav} experienced during continuous high-speed navigation in a constrained area using the proposed approach (blue) and not applying any collision avoidance mechanism (red). The upper graph shows a comparison as a histogram of normalized frequencies of minimum mutual distances. The bottom part shows a plot of the same data visualized in the form of cumulative normalized frequencies. 
  }
  \label{fig:reliability_test}
\end{figure}

Since the characteristics of the proposed algorithm do not allow for providing theoretical guarantees, we demonstrate its reliability through extensive simulations. 
In the proposed reliability test, the \acp{uav} continuously navigated to random goals in an open environment of dimensions \qtyproduct{20 x 20 x 1}{\metre} while following trajectories with an acceleration limit $a_{max} = \SI{40}{\meter\per\second\squared}$, and $r_{ca}$ set to \SI{1.0}{\meter}.
During the three-hour-long experiment, 10 \acp{uav} were navigated to more than 50000 goals, travelled a total distance of \SI[scientific-notation=true, round-mode=places,round-precision=2]{628484}{\meter} with an average velocity \SI{5.8}{\meter\per\second}, and maximum velocities up to \SI{25.9}{\meter\per\second}.
The results show that the proposed approach prevents \SI{100}{\percent} of violations of minimum mutual distance even in such a challenging scenario. 
\autoref{fig:reliability_test} presents the histogram of minimum mutual distances, quantifying the impact of the proposed method in comparison to a scenario without implemented mutual collision avoidance.

\subsection{Ablation study}
To validate the proposed approach through an ablation study, we utilize the APCX scenario with ten \acp{uav}  and a circle radius \SI{10}{\meter}.
In the study, we compare the proposed approach with the following baselines: (i) \textbf{NoTimeDep} --- proposed approach without considering time dependency of the reciprocal velocity constraints; (ii) \textbf{NoPmm} --- proposed approach with trajectory generation replaced by providing a single goal as a reference for the \ac{nmpc} controller, and (iii) \textbf{NoTDNoPmm} --- combining the absence of trajectory generation and time dependency of the reciprocal velocity constraints. 
The results show a clear benefit of the introduced time dependence of the constraints, which decreases the average flight time by $11\%$ while increasing the minimum mutual distance among \acp{uav}.
Replacing the PMM trajectory with a single goal reference lowers the flight time.
However, at the same time, it significantly decreases the safety margin and provides an unfeasible reference. This negatively affects the convergence of the defined \ac{nmpc} problem, posing a significant risk in real-world scenarios.
The detailed results of the ablation study are shown in~\autoref{tab:ablation_studies}.
\begin{table}[t]
\renewcommand{\arraystretch}{0.9}
\setlength\tabcolsep{3.0pt}
  \begin{center}
    \caption{Results of the ablation study. The presented results are averaged over 100 flights. Data shown in columns marked as \textit{min} show the minimum value of respective quantity over all trials. \textit{Min. dist.} stands for minimum mutual distance experienced between any pair of \acp{uav} during a single trial. 
    }\label{tab:ablation_studies}
    \newcommand{\rotateAngle}{00}
    \newcommand*{\OK}{\checkmark}
    \newcommand*{\hshift}{\hspace*{0.0cm}}
    \addtolength{\tabcolsep}{-0.19em} 
    \begin{tabular}{l c c c c c c c}
      \toprule 
      \multirow{2.4}{*}{Approach} & \multirow{2.4}{*}{\begin{tabular}{c}Success\\ rate [\%]\end{tabular}} & \multicolumn{2}{c}{Flight time [\si{\second}]} & \multirow{2.4}{*}{\begin{tabular}{c}Flight\\ dist. [m]\end{tabular}} & \multirow{2.4}{*}{\begin{tabular}{c}Fl. vel.\\ ~[\si{\meter\per\second}]~\end{tabular}} & \multicolumn{2}{c}{Min. dist. [\si{\meter}]}\\
 \cmidrule(){3-4} \cmidrule(){7-8}
        & & ~mean & ~min & & & ~mean & ~min \\ 
      \midrule
      proposed & \textbf{100.0} & 3.07 & 2.94 & 21.15 & 7.36 & ~\textbf{0.98} & ~\textbf{0.81} \\
      NoTimeDep & \textbf{100.0} & 3.46 & 3.02 & 21.65 & 7.19 & ~0.98 & ~0.71\\
      NoPmm  & \textbf{100.0} & \textbf{2.98} & \textbf{2.75} & \textbf{21.11} & \textbf{7.71} & ~0.90 & ~0.62\\
      NoTDNoPmm & \textbf{100.0} & 3.57 & 2.81 & 22.05 & 7.68 & ~0.89 & ~0.64\\
      \bottomrule
    \end{tabular}
    \addtolength{\tabcolsep}{+0.1em} 
  \end{center}
\end{table}

\subsection{Real-world experiments}
\label{sec:real_world_experiments}

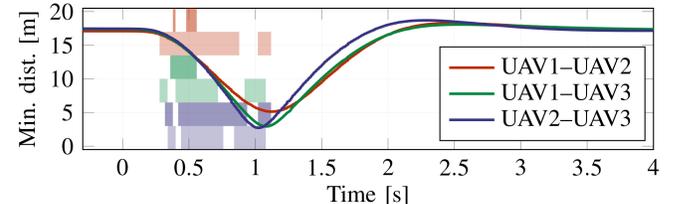
\begin{figure}[htb]
  \centering
  \definecolor{color_blue}{rgb}{0.22, 0.2, 0.502}
\definecolor{color_red}{rgb}{0.737,0.165,0}
\definecolor{color_green}{rgb}{0, .522, .243}
\hspace*{-0.22cm}

\pgfplotsset{compat=newest}
\usepgfplotslibrary{fillbetween}
\usepgfplotslibrary{groupplots}
\usetikzlibrary{intersections}

\pgfplotsset{compat=newest, 
  emphasize/.code args={#1:#2with#3}{
    \pgfplotsextra{
            \draw[color=#3, fill=#3] ({axis cs:#1,0.00} |- {axis description cs:0,0.006}) 
            rectangle ({axis cs:#2,0} |- {axis description cs:0,0.994});
    }
  }
}

\pgfplotsset{
  keep odd segments/.style={
    every segment no 0/.style={forget plot, fill=red, opacity=0.0},
    every segment no 2/.style={forget plot, fill=red, opacity=0.0},
  }
}

\pgfplotsset{
  keep even segments/.style={
    every segment no 1/.style={forget plot, opacity=0.0},
    every segment no 3/.style={forget plot, opacity=0.0},
    every segment no 5/.style={forget plot, opacity=0.0},
    every segment no 7/.style={forget plot, opacity=0.0},
    every segment no 9/.style={forget plot, opacity=0.0},
  }
}

\begin{tikzpicture}[font=\normalsize]

  \pgfplotstableread[col sep=comma]{./figs/tikz_data/output_flight_2_1.csv}{\tablepos}

  \begin{groupplot}[
    group style={
          group size=1 by 1,
          x descriptions at=edge bottom,
          vertical sep=5pt,
    },
    name=top,
    width=1.06\columnwidth,
    grid=major,
    grid style={draw=gray!12,line width=.1pt},
    xlabel= {Time [s]},
    y label style={yshift=-0.6ex},
    x label style={yshift=0.7ex},
    label style={font=\small},
    tick label style={font=\small},
    legend style={font=\small},
    ]

    \nextgroupplot[
      height=0.40\columnwidth,
      ylabel= {Min. dist. [m]},
      xmin=-0.3, xmax=4.0,
      ymin=-0.5, ymax=20.5,
      legend style={at={(0.986,0.73)}, legend columns = 1, row sep=-0.4ex},
    ]

    \addplot[color=color_red, line width=1.0pt, opacity=1.0, smooth] table[y=distance_UAV1_UAV2, x={timestamp}] {\tablepos};
    \addplot[color=color_green, line width=1.0pt, opacity=1.0, smooth] table[y=distance_UAV1_UAV3, x={timestamp}] {\tablepos};
    \addplot[color=color_blue, line width=1.0pt, opacity=1.0, smooth] table[y=distance_UAV2_UAV3, x={timestamp}] {\tablepos};


    \addplot[const plot, name path=ca12, color=color_blue, line width=0.2pt, opacity=0.0] table[x=timestamp, y=constraints_active_UAV1_on_UAV2] {\tablepos};
    \addplot[const plot, name path=ca13, color=color_red, line width=0.2pt, opacity=0.0] table[x=timestamp, y=constraints_active_UAV1_on_UAV3] {\tablepos};
    \addplot[const plot, name path=ca21, color=color_blue, line width=0.2pt, opacity=0.0] table[x=timestamp, y=constraints_active_UAV2_on_UAV1] {\tablepos};
    \addplot[const plot, name path=ca23, color=color_red, line width=0.2pt, opacity=0.0] table[x=timestamp, y=constraints_active_UAV2_on_UAV3] {\tablepos};
    \addplot[const plot, name path=ca31, color=color_blue, line width=0.2pt, opacity=0.0] table[x=timestamp, y=constraints_active_UAV3_on_UAV1] {\tablepos};
    \addplot[const plot, name path=ca32, color=color_blue, line width=0.2pt, opacity=0.0] table[x=timestamp, y=constraints_active_UAV3_on_UAV2] {\tablepos};

    \addplot[name path=bzero, red, thick, opacity=0.0] coordinates {(0,-21.5) (10,-21.5)};
    \addplot[name path=btwo, red, thick, opacity=0.0] coordinates {(0,3.0) (10,3.0)};
    \addplot[name path=bthree, red, thick, opacity=0.0] coordinates {(0,6.5) (10,6.5)};
    \addplot[name path=bfour, red, thick, opacity=0.0] coordinates {(0,10.0) (10,10.0)};
    \addplot[name path=bfive, red, thick, opacity=0.0] coordinates {(0,13.5) (10,13.5)};
    \addplot[name path=bsix, red, thick, opacity=0.0] coordinates {(0,17.0) (10,17.0)};
    \addplot[color=color_red, opacity=0.5] fill between [of=ca21 and bsix, split, soft clip={domain=0:5.0}];
    \addplot[color=color_red, opacity=0.3] fill between [of=ca31 and bfive, split, soft clip={domain=0:5.0}];
    \addplot[color=color_green, opacity=0.5] fill between [of=ca12 and bfour, split, 
    every segment no 0/.style={forget plot, opacity=0.0},
    every segment no 2/.style={forget plot, opacity=0.0},
    every segment no 4/.style={forget plot, opacity=0.0},
    soft clip={domain=0:5.0}];
    \addplot[color=color_green, opacity=0.3] fill between [of=ca32 and bthree, split, 
    every segment no 0/.style={forget plot, opacity=0.0},
    every segment no 2/.style={forget plot, opacity=0.0},
    every segment no 4/.style={forget plot, opacity=0.0},
    every segment no 6/.style={forget plot, opacity=0.0},
    soft clip={domain=0:5.0}];
    \addplot[color=color_blue, opacity=0.5] fill between [of=ca13 and btwo, split, 
    every segment no 0/.style={forget plot, opacity=0.0},
    every segment no 2/.style={forget plot, opacity=0.0},
    every segment no 4/.style={forget plot, opacity=0.0},
    every segment no 6/.style={forget plot, opacity=0.0},
    soft clip={domain=0:5.0}];
    \addplot[color=color_blue, opacity=0.3] fill between [of=ca23 and bzero, split, 
    every segment no 0/.style={forget plot, opacity=0.0},
    every segment no 2/.style={forget plot, opacity=0.0},
    every segment no 4/.style={forget plot, opacity=0.0},
    every segment no 6/.style={forget plot, opacity=0.0},
    soft clip={domain=0:5.0}];

    \addlegendentry{\small UAV1--UAV2}
    \addlegendentry{\small UAV1--UAV3}
    \addlegendentry{\small UAV2--UAV3}

  \end{groupplot}

\end{tikzpicture}
  \vspace{-0.5cm}
  \caption{Minimum distance between individual pairs of \acp{uav} in a real world flight in APCX scenario. The colored background corresponds to periods when the reciprocal velocity constraints were active on UAV1 (dark red - w.r.t. UAV2, light red - w.r.t. UAV3), UAV2 (dark green - w.r.t. UAV1, light green - w.r.t. UAV3), and UAV3 (dark blue - w.r.t. UAV1, light blue - w.r.t. UAV2).}
  \label{fig:rw_exp_min_dist}
\end{figure}

The practicality of the proposed approach was further verified through its deployment onboard UAV platforms in the APCX scenario involving three \acp{uav} with an acceleration limit up to \SI{30}{\meter\per\second\squared}, and $r_{ca} = \SI{1.5}{\meter}$ (see \autoref{fig:intro}). 
The deployed \ac{uav} platforms are based on the frame with a wheelbase of \SI{300}{\milli\meter}, equipped with RTK GPS and onboard computer, Khadas VIM3 with \SI{2}{\giga\hertz} ARM processor, running the proposed control approach along the underlying MRS UAV system \cite{baca2021mrs}.
The position and velocity of the robots were shared through a standard Wi-Fi interface with a frequency of \SI{10}{\hertz} and an average delay of \SI{14}{\milli\second}.  
The detailed results from 9 flights are presented in \autoref{tab:rw_exp} and \autoref{fig:rw_exp_min_dist}.
\begin{table}[t]
\renewcommand{\arraystretch}{0.9}
\setlength\tabcolsep{3.0pt}
  \begin{center}
    \caption{Data from real-world validation of the proposed approach. When unspecified, the mean value over individual flights is presented. \textit{Min. dist.} stands for minimum mutual distance. 
    }\label{tab:rw_exp}
    \newcommand{\rotateAngle}{00}
    \newcommand*{\OK}{\checkmark}
    \newcommand*{\hshift}{\hspace*{0.0cm}}
    \addtolength{\tabcolsep}{-0.19em} 
    \begin{tabular}{c c c c c c c c}
      \toprule 
      \multirow{2.4}{*}{\begin{tabular}{c}Ref. acc.\\ {[\si{\meter\per\second\squared}]}\end{tabular}} & \multirow{2.4}{*}{\begin{tabular}{c}Number of\\ flights [-]\end{tabular}} & \multicolumn{2}{c}{Flight time [\si{\second}]} & \multirow{2.4}{*}{\begin{tabular}{c}Flight\\ dist. [m]\end{tabular}} & \multirow{2.4}{*}{\begin{tabular}{c}Max. vel.\\ ~[\si{\meter\per\second}]~\end{tabular}} & \multicolumn{2}{c}{Min. dist. [\si{\meter}]}\\
 \cmidrule(){3-4} \cmidrule(){7-8}
        & & ~mean & ~min & & & ~mean & ~min \\ 
      \midrule
      20.0 & 2 & 5.84 & 3.94 & 23.46 & 13.88 & ~2.59 & ~1.79 \\
      30.0 & 7 & 4.18 & 3.49 & 23.35 & 17.74 & ~2.61 & ~2.20\\
      \bottomrule
    \end{tabular}
    \addtolength{\tabcolsep}{+0.1em} 
  \end{center}
  \vspace*{-0.4cm}
\end{table}

\section{Strengths and Limitations}
\label{sec:discussion}

Although the method achieves superior performance in mutual collision avoidance for both low- and high-speed scenarios, its limitations include the assumption of an obstacle-free environment, lack of theoretical guarantees, and the need for frequent updates of other robots’ states.
The integration of obstacles, such as spheres, cylinders, or planes, is straightforward, as they can be modelled as velocity obstacles. However, general-shaped obstacles require deeper analysis being a subject of the future work.

Despite the absence of theoretical guarantees, the method’s ability to prevent collisions is demonstrated in simulations under realistic conditions.
During three hours of testing, no collisions occurred, though a significant drop in the mutual distance of \acp{uav} was observed a few times.
These drops were caused by highly dynamic scenarios with asynchronous goal changes in a small environment.
The most critical cases occur when new velocity constraints are activated while an \ac{uav} is already at maximum acceleration to avoid another \ac{uav} in its vicinity.
However, such cases are extremely rare and would occur only in environments with UAV densities exceeding those expected in real-world applications.

The need for frequent updates of other \acp{uav}' states arises from sharing only limited information.
With high acceleration limits, the prediction of other \acp{uav}' states, based on the last received position and velocity, and a first-order linear motion model, can significantly diverge from the real state within just tens of milliseconds.
To preserve avoidance capabilities, uncertainty resulting from message delays, low update rates, and other factors must be accounted for in the setting of the reference collision-avoidance radius.

Alongside the small amount of data that is required to avoid collisions, a key advantage of the proposed method lies in the independence between the length of the \ac{nmpc} control horizon and the horizon used for detecting potential collisions and generating velocity constraints. 
Thus, allowing avoidance maneuvers to be initiated several seconds in advance while keeping the \ac{nmpc} horizon short to maintain real-time performance.
As a result, the maneuvers are smoother and the method remains practical for vehicles with lower maneuverability, and even  under message delays longer than the \ac{nmpc} horizon.



\section{Acknowledgement}
\label{sec:acknowledgement}
The authors would like to thank Manuel Boldrer and Charbel Toumieh for their help with tuning their methods for the purpose of the presented comparison in the APCX scenario. 


\bibliographystyle{IEEEtran}
\bibliography{main.bib}

\end{document}